%% file: main.tex
\documentclass[letterpaper]{article} 
\usepackage{aaai2026}  
\usepackage{times}  
\usepackage{helvet}  
\usepackage{courier}  
\usepackage[hyphens]{url}  
\usepackage{graphicx} 
\usepackage{natbib}  
\usepackage{caption} 
\usepackage{algorithm}
\usepackage{algorithmic}

\usepackage{newfloat}
\usepackage{listings}
\DeclareCaptionStyle{ruled}{labelfont=normalfont,labelsep=colon,strut=off} 
\floatstyle{ruled}
\newfloat{listing}{tb}{lst}{}
\floatname{listing}{Listing}
\usepackage{amsfonts}       
\usepackage{nicefrac}       

\usepackage{booktabs}       
\usepackage{makecell}       
\usepackage{array}          
\usepackage{multirow}       
\usepackage{adjustbox}      

\usepackage{microtype}      
\usepackage{xcolor}         
\usepackage{amsmath}
\usepackage{enumitem}

\usepackage{times}  
\usepackage{helvet}  
\usepackage{courier}  
\usepackage[hyphens]{url}  
\usepackage{graphicx} 
\usepackage{natbib}  
\usepackage{caption} 
\usepackage{algorithm}
\usepackage{algorithmic}
\usepackage{times}
\usepackage{helvet}
\usepackage{courier}
\usepackage{xcolor}

\usepackage{newfloat}
\usepackage{listings}
\DeclareCaptionStyle{ruled}{labelfont=normalfont,labelsep=colon,strut=off} 
\floatstyle{ruled}
\newfloat{listing}{tb}{lst}{}
\floatname{listing}{Listing}

\title{GeoX-Bench: Benchmarking Cross-View Geo-Localization and Pose Estimation Capabilities of Large Multimodal Models}

\author{
    Yushuo Zheng\textsuperscript{\rm 1,2},
    Jiangyong Ying\textsuperscript{\rm 3},
    Huiyu Duan\textsuperscript{\rm 1}\footnotemark[1],
    Chunyi Li\textsuperscript{\rm 1,2},
    Zicheng Zhang\textsuperscript{\rm 1,2},
    Jing Liu\textsuperscript{\rm 4},
    Xiaohong Liu\textsuperscript{\rm 1,5}\footnotemark[1],
    Guangtao Zhai\textsuperscript{\rm 1,2}\thanks{Corresponding author.}
}

\affiliations{
    \textsuperscript{\rm 1}Shanghai Jiao Tong University \quad
    \textsuperscript{\rm 2}Shanghai AI Lab \quad
    \textsuperscript{\rm 3}China Telecom \\
    \textsuperscript{\rm 4}Tianjin University \quad
    \textsuperscript{\rm 5}Shanghai Jiao Tong University Sichuan Research Institute \\
    }

\begin{document}

\maketitle

\begin{abstract}

Large multimodal models (LMMs) have demonstrated remarkable capabilities across a wide range of tasks, however their knowledge and abilities in the cross-view geo-localization and pose estimation domains remain unexplored, despite potential benefits for navigation, autonomous driving, outdoor robotics, \textit{etc}.
To bridge this gap, we introduce \textbf{GeoX-Bench}, a comprehensive \underline{Bench}mark designed to explore and evaluate the capabilities of LMMs in \underline{cross}-view \underline{Geo}-localization and pose estimation.
Specifically, GeoX-Bench contains 10,859 panoramic-satellite image pairs spanning 128 cities in 49 countries, along with corresponding 755,976 question-answering (QA) pairs. Among these, 42,900 QA pairs are designated for benchmarking, while the remaining are intended to enhance the capabilities of LMMs.
Based on GeoX-Bench, we evaluate the capabilities of 25 state-of-the-art LMMs on cross-view geo-localization and pose estimation tasks, and further explore the empowered capabilities of instruction-tuning. Our benchmark
demonstrate that while current LMMs achieve impressive performance in geo-localization tasks, their effectiveness declines significantly on the more complex pose estimation tasks, highlighting a critical area for future improvement, and instruction-tuning LMMs on the training data of GeoX-Bench can significantly improve the cross-view geo-sense abilities. The GeoX-Bench is available at 
\textcolor{magenta}{https://github.com/IntMeGroup/GeoX-Bench}.
\end{abstract}

\input{secs/1-intro}

\input{secs/2-related}
\input{secs/3-Geo-Bench}
\input{secs/4-Exp}
\input{secs/5-Conclusion.tex}
\input{secs/7-ack}

\newpage
\bibliography{references}
\end{document}

%% file: secs/1-intro.tex
\section{Introduction}

Accurately determining the geographic location and pose of the camera is a fundamental challenge in computer vision and robotics, with broad implications for applications such as autonomous driving, robotic navigation, and embodied AI agents operating in real-world environments. These scenarios often involve degraded or unavailable GPS signals and magnetic interference, necessitating robust visual-based geospatial reasoning, \textit{i.e.}, cross-view geo-localization and pose estimation. 
However, the inherent complexity of the physical world, manifested in its large-scale spatial layout and diverse visual appearance, makes this task particularly difficult. Recent advances in large multimodal models (LMMs), such as GPT-4V \cite{openai2024gpt4technicalreport} and LLaVA \cite{liu2023llava} have demonstrated impressive capabilities in free-form visual reasoning across a range of tasks. Despite this progress, their proficiency in geospatial grounding, the ability to understand and reason about geographic location and spatial orientation, remains largely unexamined, due to the absence of standardized, large-scale benchmarks specifically designed to assess such capabilities.

Existing efforts to evaluate the geographic understanding of LMMs have primarily focused on satellite image geo-localization, which provides only coarse-grained location information and lacks the estimation of camera pose \cite{Hu_2018_CVPR}. Another task, geo-guessing, requires models to predict the approximate location of a ground-level image only based on its visual content \cite{Weyand_2016}. However, this task does not involve the cross-view geo-localization and pose estimation problem, which aims to determine both the geographic location and pose based on the captured ground-level image and the aerial perspective image. This more comprehensive task demands a deeper and more structured understanding of spatial relationships in the visual scene, presenting a more rigorous test of LMMs' geospatial reasoning capabilities.

\begin{figure*}[t]
    \centering
    \includegraphics[width=\textwidth]{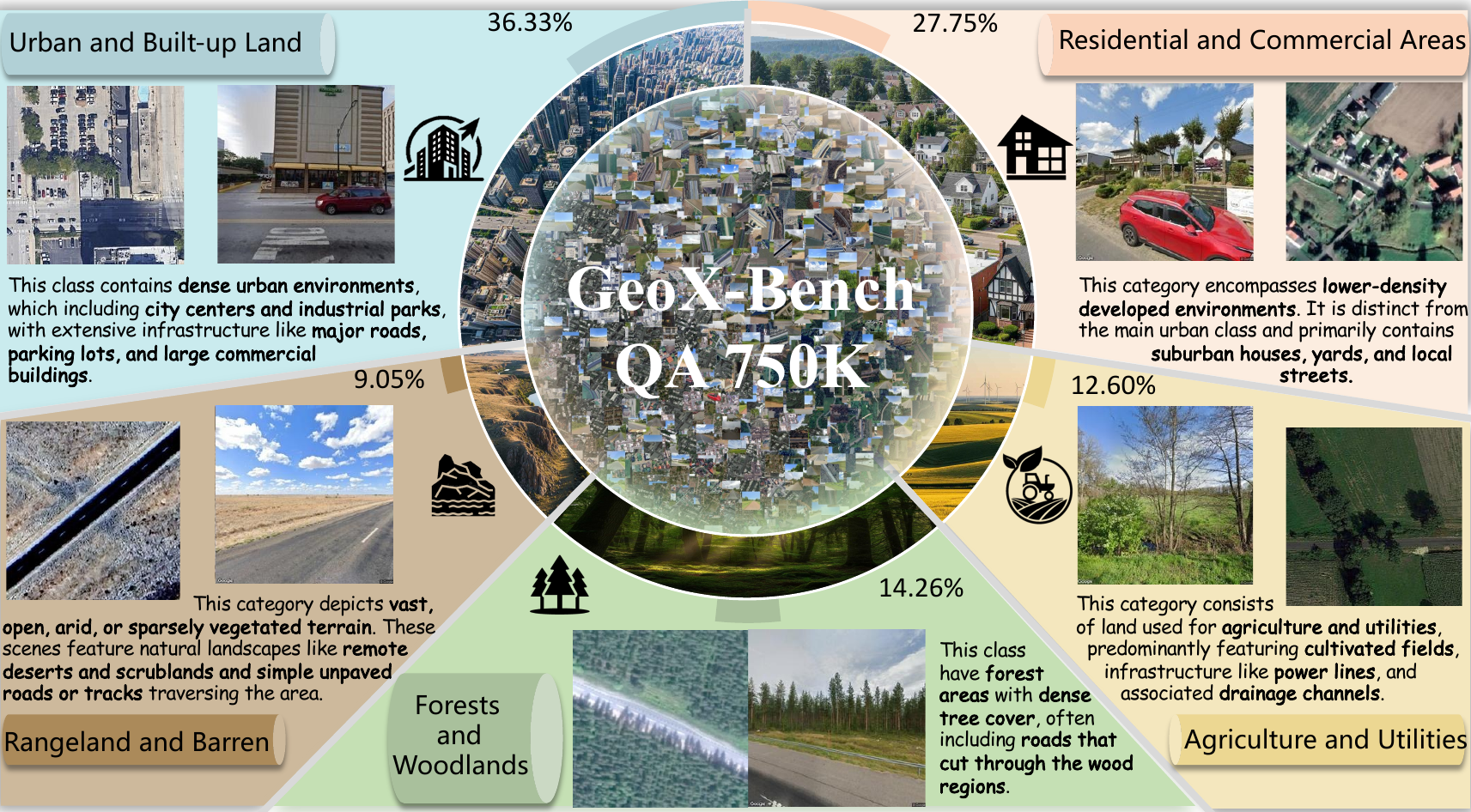}
    \caption{Geographic composition of the \textbf{GeoX-Bench} dataset by land cover type. The benchmark is weighted toward developed regions, with Urban/Built-up (36.33\%) and Residential/Commercial (27.75\%) areas constituting the majority. Natural and rural environments, including Forests (14.26\%), Agriculture (12.60\%), and Rangeland/Barren (9.05\%), provide a geospatially representative representative diverse settings for evaluation.}
    \label{fig:LandType}
\end{figure*}

In this paper, we introduce a new benchmark termed \textbf{GeoX-Bench} for evaluating the capabilities of LMMs on cross-view geo-localization and pose estimation tasks.GeoX-Bench comprises 10,859 panoramic-satellite image pairs covering 128 cities across 49 countries, along with 755,976 carefully curated corresponding question-answering (QA) pairs. Among these, 42,900 QA pairs are curated for standardized benchmarking, while the remaining are intended to enhance model capabilities through instruction tuning.
GeoX-Bench defines two core tasks,   including \textbf{(1) geo-localization}, \textit{i.e.}, predicting the geographic location of a ground-level perspective image given a satellite image; and \textbf{(2) geo-pose estimation}, \textit{i.e.}, inferring the camera’s orientation of a ground-level perspective image given a satellite image. 

Based on GeoX-Bench, we establish rigorous evaluation protocols, and evaluate the capabilities of 25 state-of-the-art LMMs and instruction-tuned LMMs on cross-view geo-localization and pose estimation tasks. Our experimental results reveal that while contemporary LMMs achieve promising performance on geo-localization, their capabilities degrade considerably on the more challenging pose estimation task. Moreover, the significant improvement brought by the instruaction-tuning suggests that the capabilities of LMMs on both cross-view geo-localization and pose estimation tasks can be further improved.

Our main contributions are as follows:
\begin{itemize}
    \item We introduce GeoX-Bench, a large-scale benchmark comprising 10,859 panoramic-satellite image pairs and 755,976 curated QA pairs designed for evaluating LMMs' capabilities on cross-view geo-localization and pose estimation.
    \item We propose rigorous evaluation metrics for both tasks and benchmark 25 state-of-the-art LMMs and instruction-tuned variants.
    \item We provide a comprehensive analysis of the performance of LMMs on both tasks, highlighting the strengths and weaknesses of current models and identifying critical areas for future improvement.
\end{itemize}

%% file: secs/2-related.tex
\begin{figure*}[t]
    \centering
    \includegraphics[width=\linewidth]{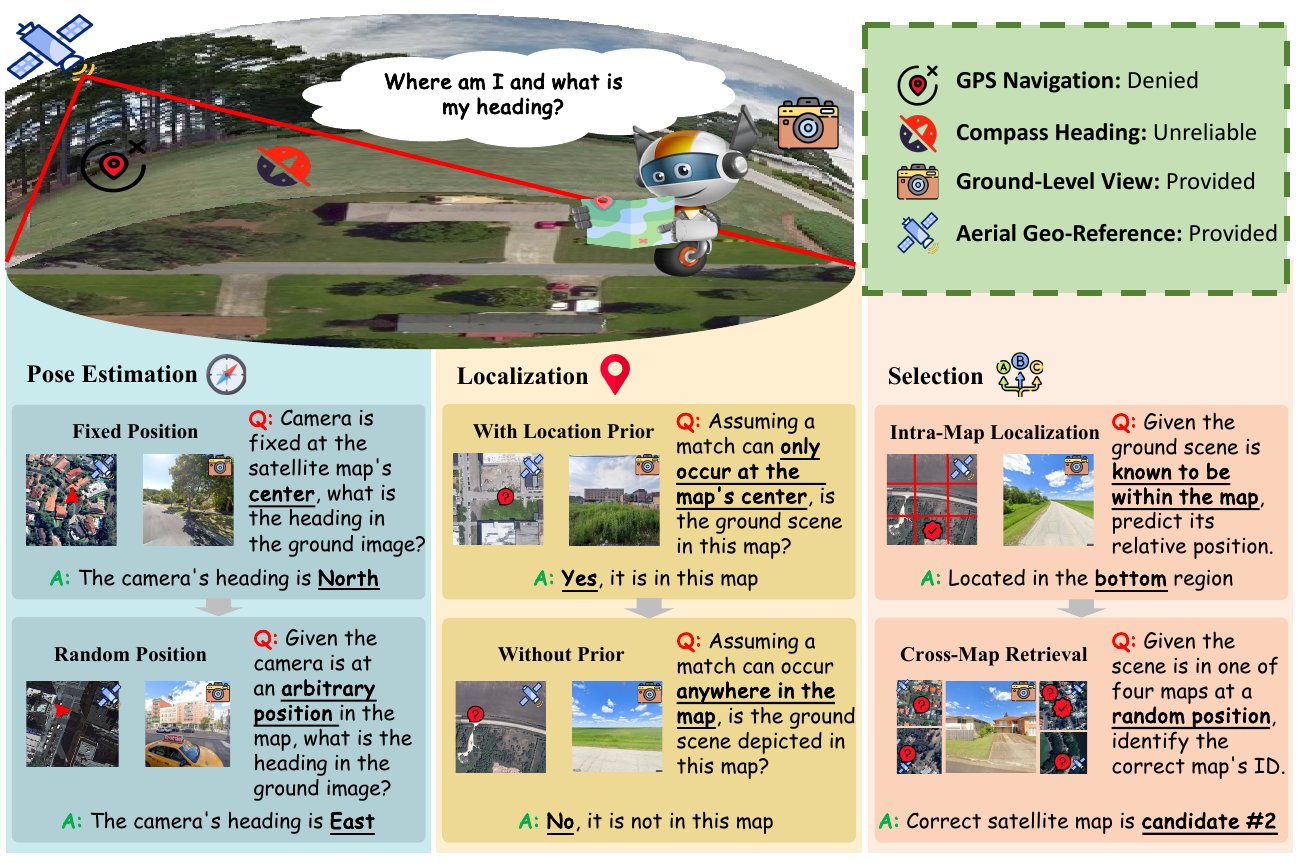}
    \caption{Illustration of the \textbf{GeoX-Bench} benchmark tasks. The tasks include heading estimation with known or unknown camera locations, location verification on a satellite map, location selection, and map selection from candidates. These tasks evaluate models' abilities to reason over ground-to-satellite image pairs for localization and pose understanding.}
    \label{fig:Task}
\end{figure*}

\section{Related Work}

\subsection{Large Multimodal Models and Their Spatial Reasoning Capability}
The rapid evolution from Large Language Models (LLMs) \cite{brown2020languagemodelsfewshotlearners} to Large Multimodal Models (LMMs) such as LLaVA \cite{liu2023llava}, GPT-4 \cite{openai2024gpt4technicalreport}, and Gemini \cite{G1d5} has unlocked advanced capabilities in processing both text and vision \cite{Duan_2025_CVPR,duan2024uniprocessor,10.1145/3746027.3755405,yang2025odibenchmllmsunderstandimmersive}. LMMs have been increasingly applied to embodied intelligence \cite{mai2023llmroboticbrainunifying}, where spatial awareness is essential \cite{jin2025embodiedworldmodelsemerge}. Early explorations, such as LLMGeo \cite{wang2024llmgeo} for coarse location reasoning, LLaVA-3D \cite{zhu2024llava} for pose understanding, and All-Angles Bench \cite{yeh2025seeingperspectiveevaluatingmultiview}, demonstrate emerging capabilities, while recent competitive benchmarks \cite{zheng2025lmfightarenabenchmarking} broaden evaluation perspectives. However, existing studies address spatial skills in isolation and lack the precision required for real-world applications. To address this, our work proposes fine-grained cross-view localization and pose-estimation tasks that more directly evaluate integrated spatial reasoning.

\subsection{Cross-View Geo-Localization and Geo-Pose Estimation}
Evaluation methods have evolved alongside vision–language models. Foundational benchmarks such as VQA \cite{Goyal2016MakingTV} established multimodal reasoning, while recent works like MME \cite{fu2023mme} and MMMU \cite{yue2024mmmu} assess complex, expert-level tasks. Broader evaluation platforms \cite{aibench} and perceptual studies \cite{chen2025just} further diversify assessment frameworks. In remote sensing, benchmarks like RSVQA \cite{lobry2020rsvqa} and GeoChat \cite{kuckreja2024geochat} focus primarily on aerial imagery, while immersive-environment benchmarks \cite{ji2025medomni,ji2025evaluating,ji2025assessing,ji2024application} assess medical or panoramic understanding. Despite these efforts, a gap persists: systematic evaluation of ground-to-satellite spatial reasoning. Existing cross-view datasets—CVUSA \cite{CVUSA}, VIGOR \cite{zhu2021vigor}, OmniCity \cite{li2022omnicity}, LLMGeo \cite{wang2024llmgeo}, CVGlobal \cite{CVGloable}—provide strong foundations yet remain fragmented and geographically biased. \textbf{GeoX-Bench} unifies six diverse datasets to create the first geographically broad, multi-view benchmark designed specifically to assess both fine-grained geo-localization and orientation reasoning, as illustrated in Figure~\ref{fig:LandType}.

%% file: secs/3-Geo-Bench.tex
\section{GeoX-Bench}
\begin{figure*}[t]
    \centering
    \includegraphics[width=\linewidth]{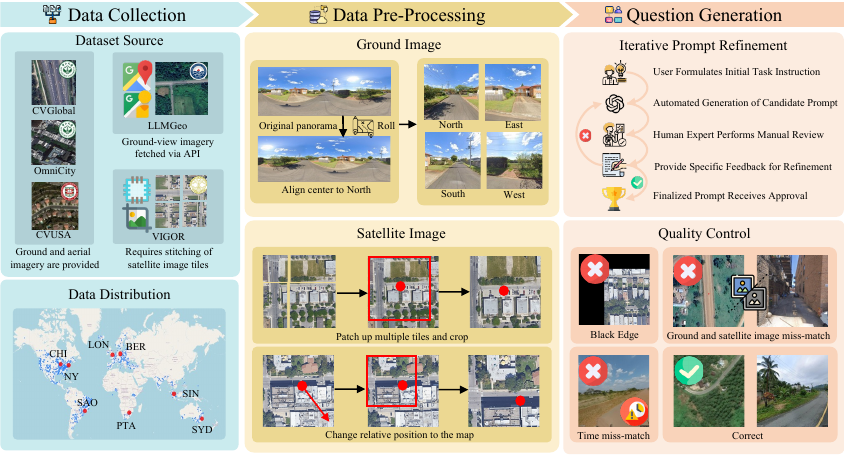}
    \caption{The \textbf{GeoX-Bench} data curation pipeline, from source sampling to final quality control. We first sample ground-satellite pairs from four existing datasets to ensure broad geographic coverage. In the pre-processing stage, ground-level panoramas are programmatically rotated to align to a consistent North orientation before cardinal views are extracted, while corresponding satellite imagery is stitched and cropped. An iterative, LLM-assisted framework with human oversight is used for question prompt generation. A final quality control stage removes data with visual artifacts or cross-view inconsistencies, such as spatial or temporal mismatches, to ensure benchmark integrity.}
    \label{fig:DataPipeline}
\end{figure*}

\subsection{Overview}
We propose \textbf{GeoX-Bench}, a novel benchmark specifically designed for evaluating the capability of Large Multimodal Models (LMMs) in understanding geo-localization and pose estimation in outdoor environments. \textbf{GeoX-Bench} comprises a training set of 10,684 panorama-satellite pairs and a test set of 175 pairs, distributed across the world as shown in Figure \ref{fig:LandType}. This corresponds to a total of over 750,000 question-answer pairs, systematically divided into training and test sets while maintaining a consistent question and answer distribution. For detailed statistical information, please refer to the supplemental material. GeoX-Bench is characterized by: (1) comprehensive cross-view and multimodal information, where each image pair or group includes at least one ground-level image and one satellite image; and (2) extensive variability in land types, facilitating the assessment of LMM performance across diverse scenarios. GeoX-Bench encompasses seven specific tasks, covering both \textbf{geo-localization} and \textbf{geo-pose estimation} as shown in Figure \ref{fig:Task}.

\subsubsection{Comparison with Existing Benchmarks}
While general benchmarks such as MMBench \cite{MMBench} effectively assess the overall performance of LMMs, they predominantly evaluate single-view capabilities. In contrast, \textbf{GeoX-Bench} uniquely enables the evaluation of LMMs across different viewpoints and scenarios. Benchmarks like LLMGeo \cite{wang2024llmgeo} assess model capabilities in geographical location inference but are limited to single-view inputs, lacking precise geo-localization accuracy and neglecting pose estimation capabilities. Additionally, benchmarks such as All-Angles Bench \cite{yeh2025seeingperspectiveevaluatingmultiview}, though designed to evaluate multi-view understanding, differ fundamentally from GeoX-Bench by providing multiple views from a single modality. Conversely, GeoX-Bench distinguishes itself by incorporating both ground-level and satellite imagery, thereby enabling robust multi-modal and cross-view evaluation.

\input{tables/BenchTable}
\subsection{Benchmark Tasks}

\subsubsection{Pose estimation with Fixed Camera Position} Keeps the ground camera at the center of a north-aligned satellite image and asks the model to infer the absolute heading of the camera from North, East, South and West.
\subsubsection{Pose estimation with Random Camera Position} Put the camera to an random position within the map while preserving the north-aligned prior of the satellite image and asks the model to infer the absolute heading of the camera from North, East, South and West.

\subsubsection{Localization with Location Prior} The ground-view camera is assumed fixed at the geometric center of the correct north-oriented satellite image. The model must then issue a binary decision on whether the scene depicted in the ground image lies within that image.

\subsubsection{Localization without Location Prior} This task lifts the center constraint. The camera can be anywhere inside the satellite image, and the model must decide if the ground image is taken from that area.

\subsubsection{Intra-Map Localization} Assuming the ground image is contained within the map randomly. The model is then asked to predict its relative position with respect to the map center.

\subsubsection{Standard Cross-Map Retrieval} The ground viewpoint is centered in its correct satellite image. A model is given the ground photograph along with four candidate maps and must return the index of the matching map.

\subsubsection{Random Cross-Map Retrieval} This task repeats the retrieval procedure, but with the camera placed at an arbitrary position inside the correct map. This forces the model to rely on holistic scene understanding without location prior.

\subsection{Benchmark Curation}

\subsubsection{Data Collection}
As illustrated in Figure \ref{fig:DataPipeline}, the GeoX-Bench dataset was created by sampling data from several established datasets. To ascertain model robustness, we sampled a balanced number of data points from the CVGlobal, CVUSA, OmniCity, LLMGeo, and VIGOR datasets. Since temporal synchronization between satellite and ground-level imagery is crucial, we adopted methodologies from prior research. Specifically, we re-accessed the Google Maps API to obtain current satellite imagery corresponding to the street-view images in the datasets, ensuring both views were captured between 2024 and 2025.

\subsubsection{Data Pre-processing}
For ground-level images, panoramic views from datasets like CVGlobal and OmniCity were first rotated to a standard North-facing orientation. Following prior methodology \cite{wang2024llmgeo}, each panorama was then transformed into four images with a 90-degree Field of View (FoV), each oriented towards a cardinal direction (North, East, South, and West). For aerial images, particularly from the VIGOR dataset where satellite and ground views do not have a one-to-one correspondence, we merged and resized satellite imagery to a uniform 512×512 pixels. This process ensures the ground-level perspective is precisely centered within its corresponding satellite image. For map randomization tasks, we used off-center cropping to reposition these central locations arbitrarily.

\subsubsection{Prompt Generation}
We used GPT-4o for prompt generation. Each prompt consists of a system prompt and a user prompt. The system prompt describes the input data, defines the task, and concludes with explicit formatting instructions and examples. The user prompt provides the relevant images, reiterates the task, and reinforces the required output format. After generation, all prompts were manually reviewed to ensure quality and adherence to requirements.

\subsubsection{Quality Control}
Samples exhibiting visual artifacts—such as black padding, corrupted tiles, or missing imagery—were immediately removed. The remaining pairs were then scrutinized for cross-view and temporal coherence. Each street-level photograph had to spatially align with its satellite counterpart, and the capture dates had to be close enough to prevent significant appearance changes. Any instances with clear mismatches (e.g., an urban alley paired with a rural tile) or large time gaps were discarded.

%% file: tables/BenchTable.tex
\begin{table*}[t]
\centering
\scriptsize
\resizebox{0.80\textwidth}{!}{
\setlength{\tabcolsep}{1mm}
\begin{tabular}{c|c|cc|cc|ccc|c}
\toprule
\multirow{2}{*}{\textbf{Type}} & \multirow{2}{*}{\textbf{Model}} & \multicolumn{2}{c|}{\textbf{Pose Estimation}} & \multicolumn{2}{c|}{\textbf{Localization}} & \multicolumn{3}{c|}{\textbf{Intra \& Cross map Selection}} & \multirow{2}{*}{\textbf{Average}} \\
\cmidrule{3-9}
& & \textbf{Fixed} & \textbf{Random} & \textbf{w/ Prior} & \textbf{w/o Prior} & \textbf{\makecell{IM Rnd.}} & \textbf{\makecell{CM Std.}} & \textbf{\makecell{CM Rnd.}} & \\
\midrule
\textbf{Random} & Random Selection & 25.00 & 25.00 & 50.00 & 50.00 & 11.11 & 25.00 & 25.00 & 30.16 \\
\midrule
\multirow{21}{*}{\textbf{\makecell{Open\\Source}}} & DeepSeek-VL-7B & 25.36 & 25.07 & 50.64 & 51.48 & 11.63 & 19.91 & 22.51 & 29.51 \\
& InternVL2-2B & 24.00 & 24.38 & 50.09 & 49.79 & 10.02 & 25.73 & 25.74 & 29.96 \\
& InternVL2-4B & 24.45 & 24.45 & 50.18 & 50.31 & 11.92 & 24.18 & 24.12 & 29.95 \\
& InternVL2-8B & 26.27 & 24.86 & 62.09 & 60.73 & 11.30 & 30.64 & 28.33 & 34.89 \\
& InternVL2-40B & 25.09 & 25.04 & 70.09 & 66.63 & 11.33 & 29.91 & 28.75 & 36.69 \\
& InternVL2-76B & 25.27 & 25.21 & 63.18 & 62.43 & 10.22 & 34.91 & 33.61 & 36.41 \\
& InternVL3-2B & 25.73 & 25.75 & 57.09 & 58.48 & 11.41 & 28.27 & 27.03 & 33.40 \\
& InternVL3-8B & 25.36 & 25.12 & 68.36 & 65.41 & 11.47 & 43.64 & 41.21 & 40.08 \\
& InternVL3-38B & 27.64 & 25.09 & 76.27 & 71.86 & 6.67 & 69.73 & 64.82 & 48.87 \\
& InternVL3-78B & 27.64 & 27.89 & 74.64 & 70.24 & 13.29 & 68.45 & 61.45 & 49.09 \\
& LLaVA-Interleave-7B & 25.73 & 24.79 & 51.27 & 52.09 & 11.41 & 24.00 & 24.46 & 30.54 \\
& LLaVA-One-Vision-7B & 22.82 & 24.51 & 58.91 & 56.76 & 11.42 & 26.64 & 27.81 & 32.69 \\
& mPLUG-Owl3-7B & 24.00 & 18.85 & 53.82 & 53.16 & 10.41 & 19.00 & 19.03 & 28.32 \\
& MS-Phi3d5 & 24.18 & 24.21 & 54.27 & 52.27 & 11.07 & 21.82 & 22.93 & 30.11 \\
& Qwen2.5VL-3B & 27.73 & 25.54 & 50.00 & 49.81 & 11.46 & 24.73 & 24.89 & 30.59 \\
& Qwen2.5VL-7B & 23.27 & 24.99 & 65.00 & 63.10 & 11.26 & 44.00 & 38.30 & 38.56 \\
& Qwen2.5VL-32B & 22.27 & 19.91 & 72.27 & 70.87 & 11.69 & 56.82 & 52.53 & 43.76 \\
& Qwen2.5VL-72B & 29.91 & 27.53 & 71.73 & 68.69 & 12.73 & 65.55 & 61.98 & 48.30 \\
& Qwen2VL-2B & 24.91 & 24.84 & 47.91 & 50.42 & 11.18 & 24.91 & 27.11 & 30.18 \\
& Qwen2VL-7B & 25.00 & 24.16 & 64.73 & 61.15 & 11.99 & 32.09 & 33.07 & 36.03 \\
& Qwen2VL-72B & 27.55 & 26.71 & 71.18 & 66.79 & 11.76 & 70.27 & 60.28 & 47.79 \\
\midrule
\multirow{4}{*}{\textbf{\makecell{Closed\\Source}}} & Claude-Sonnet-4 & 29.55 & 37.12 & 68.18 & 62.88 & 14.39 & 43.94 & 56.82 & 44.70 \\
& Gemini-2.5-Pro & 45.45 & \underline{41.67} & 87.88 & 79.55 & 18.94 & 85.61 & 77.27 & 62.34 \\
& GPT-4o & 33.33 & 26.52 & 90.15 & 82.58 & 15.15 & 80.30 & 81.06 & 58.44 \\
& o3 & 50.00 & \textbf{46.97} & 84.09 & 81.06 & 18.94 & 84.09 & 78.79 & 63.42 \\
\midrule
\multirow{4}{*}{\textbf{\makecell{Instruaction\\Tuning\\(LoRA)}}} & InternVL2-8B & 25.00 & 24.87 & \textbf{98.82} & 84.82 & 24.58 & \underline{87.45} & 78.73 & 60.61 \\
& InternVL3-8B & \textbf{55.36} & 40.80 & \underline{98.45} & \textbf{91.97} & \textbf{36.33} & \textbf{91.73} & \textbf{84.80} & \textbf{71.35} \\
& Qwen2.5VL-7B & 51.36 & 36.87 & 98.09 & 81.08 & \underline{35.80} & 88.55 & 79.94 & 67.38 \\
& Qwen2VL-7B & \underline{53.27} & 38.73 & 98.27 & \underline{82.85} & 35.62 & 90.27 & \underline{83.71} & \underline{68.96} \\
\bottomrule
\end{tabular}
}
\caption{Performance comparison of various vision-language models across different tasks, organized by model type. The highest score in each column is shown in \textbf{bold}, and the second-highest is \underline{underlined}. The IM Rnd. means Intra-Map Localization with random location, CM Std. means Cross-Map Retrieval with location at center of the map prior and CM Rnd. means Cross-Map Retrieval without location prior.}
\label{tab:model-evaluation}
\end{table*}

%% file: secs/4-Exp.tex
\section{Experiments}
\subsection{Evaluation Setup}
 Our analysis includes 21 open-source models and 4 leading closed-source models to provide a comprehensive comparison. The open-source models tested cover several prominent families, including the InternVL series \cite{chen2024internvl, internvl3}, the QwenVL series \cite{Qwen2-VL, Qwen2.5-VL}, and models from the LLaVA family \cite{li2024llavanextinterleavetacklingmultiimagevideo, li2024llavaonevisioneasyvisualtask}. We also include other notable models such as mPLUG-Owl3-7B \cite{ye2024mplugowl3longimagesequenceunderstanding} and Phi-3.5-Vision-Instruct \cite{abdin2024phi3technicalreporthighly}.  The close-source models including Claude-Sonnet-4 \cite{TheC3}, Gemini-2.5-Pro \cite{comanici2025gemini25pushingfrontier}, GPT-4o \cite{openai2024gpt4technicalreport} and o3 \cite{o3}.For additional context, we include benchmarks of four instruction turned model on a portion of the GeoX-Bench training dataset. To ensure reproducibility, we set the temperature to 0 and per-form greedy decoding.

\input{tables/Bias}

\begin{figure}[t]
  \centering
  \includegraphics[width=\linewidth]{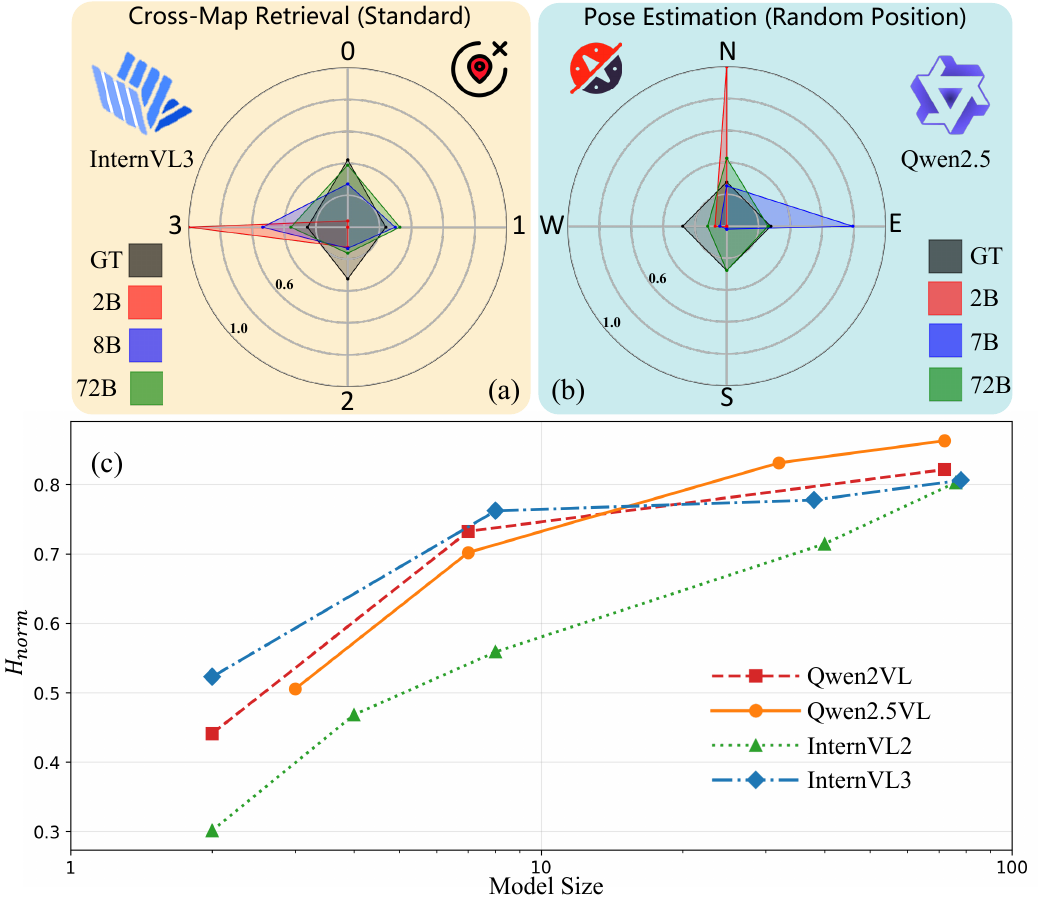}
  \caption{Model scale mitigates choice bias across different tasks and model families. \textbf{(a, b)} The min-max normalized radar charts show that smaller models exhibit a strong preference for a single option on both the \textit{Cross-Map Retrieval} and \textit{Fixed-Pose Heading Estimation} tasks. This bias diminishes as model size increases, resulting in more uniform choice distributions. \textbf{(c)} This trend is generalized across four model families, showing that normalized entropy ($H_{\text{norm}}$) consistently increases with the number of parameters. This demonstrates that larger models are better calibrated and less prone to relying on simplistic choice priors.}
  \label{fig:entropy_vs_size}
\end{figure}

To investigate the benefits of task-specific training, we conduct instruction turning on representative models from the Qwen-VL and InternVL families. Our finetuning dataset is a curated subset of the GeoX-Bench training data, containing 31,392 question-answer pairs derived from 654 unique locations. The training is focused exclusively on the three non-randomized tasks: Pose Estimation (Fixed), Localization with Prior and Cross-Map Selection. And the Intra-Map Localization task. 

We employ the parameter-efficient Low-Rank Adaptation (LoRA) \cite{hu2022lora} technique for all finetuning experiments. The models are trained for a total of 4 epochs on NVIDIA A800 GPU, utilizing the SWIFT \cite{zhao2024swiftascalablelightweightinfrastructure} framework. During training, the vision transformer (ViT) backbone remains frozen to preserve its learned general-purpose visual features. We adopt a bfloat16 mixed-precision setup and leverage Flash Attention \cite{dao2023flashattention2} to optimize computational efficiency. 

Key hyperparameters are kept consistent across all instruction turning runs: a learning rate of $1 \times 10^{-4}$, a per-device batch size of 2, and a warmup ratio of 0.05. For LoRA, we configure the rank ($r$) to 8, the scaling factor ($\alpha$) to 32, and apply the adaptation to all linear layers. The total training time is approximately 20 hours for QwenVL models and 30 hours for InternVL models. For final evaluation, we select the checkpoint that achieves the highest word accuracy on the validation set that split from the  instruction turning data.

\subsection{Main Results}

\subsubsection{LMM Performance Trends}
The evluation data across all 7 tasks are shown in Table \ref{tab:model-evaluation}, which reveals three key insight as following.(\textbf{a})Task difficulty: Geo-localization is significantly easier than pose estimation. Even 2B parameter models outperform random chance on localization tasks, whereas no model surpasses 30\% accuracy on heading estimation without fine-tuning.(\textbf{b})Scaling laws: Accuracy generally increases with model scale up to around 32B parameters, after which performance tends to saturate for pre-trained models.(\textbf{c})Fine-tuning impact: Instruction turning drastically improves performance. Notably, the \textbf{InternVL3-8B} model, when fine-tuned on our training set, outperforms all other models, including the much larger 72B parameter variants, demonstrating the high value of task-specific data.

\subsubsection{Option Preference of LMMs}

To analyze how Large Language Models (LMMs) distribute their choices, we employ several metrics. First, to quantify the dispersion of selections across a fixed set of choices, we define the empirical probability $p_i$ for each of the $k$ possible outcomes based on the model's raw selection counts $n_i$:
\[
p_i = \frac{n_i}{\sum_{j=1}^k n_j}
\]

To measure the uniformity of this choice distribution, we employ \textbf{normalized entropy}. The Shannon entropy $H$ of this discrete distribution is given by $H = - \sum_{i=1}^k p_i \log_2 p_i$. To make this metric comparable across tasks with a different number of choices $k$, we normalize $H$ by the maximum possible entropy, $\log_2 k$. The resulting metric, $H_{\mathrm{norm}}$, ranges from 0 (complete concentration on a single choice) to 1 (a perfectly uniform distribution):
\[
H_{\mathrm{norm}} = \frac{H}{\log_2 k} = -\frac{1}{\log_2 k} \sum_{i=1}^k p_i \log_2 p_i
\]
In our experiments, we observe a clear trend: as a model’s size increases, its normalized entropy $H_{\mathrm{norm}}$ also increases. This indicates that \emph{larger models tend to distribute their selections more uniformly}. As shown in Figure \ref{fig:entropy_vs_size}, the positive correlation underscores that bigger models exhibit higher normalized entropy and thus less extreme preference for any single choice.

To directly measure the strength of a model's preference for its single favorite option, we define the \textbf{mode probability}, $p_{\text{mode}}$, as the empirical probability of the most frequently selected choice, $k^\star$:
\[
p_{\text{mode}} = \max_k p_k
\]
A high $p_{\text{mode}}$ signifies a strong concentration on one choice. We can evaluate the impact of this primary bias by calculating the \textbf{mode-only baseline} accuracy, $a_{\text{mode}}$, which is the accuracy achieved if the model only ever predicted this single choice $k^\star$:
\[
a_{\text{mode}} = \frac{1}{N}\sum_{i=1}^N \mathbf{1}[y_i = k^\star]
\]
where $y_i$ is the ground-truth answer. Finally, to quantify the overall deviation of a model's answer distribution $\mathbf{p}$ from a uniform random guess $\mathbf{u}$, we use the \textbf{bias magnitude} as captured by the Kullback-Leibler (KL) divergence:
\[
D_{\mathrm{KL}}(\mathbf{p} \parallel \mathbf{u}) = \sum_{i=1}^{K} p_{i} \log_2\frac{p_{i}}{1/K}
\]

\subsubsection{Empirical Findings}

The data in Table \ref{tab:option-bias} reveals a clear pattern: \textbf{smaller models exhibit a strong choice bias that diminishes as model scale increases}. Models in the 2B to 4B parameter range consistently display a pronounced preference for a single option. For example, Qwen2VL-2B assigns over 73\% of its predictions to its most-favored choice ($p_{\text{mode}} = 0.7342$), resulting in a high KL divergence from uniform ($D_{\text{KL}} = 1.0985$ bits) and a low normalized entropy ($H_{\text{norm}} = 0.4412$).

Notably, this predictive bias is a consistent in a model. The accuracy of a \textbf{mode-only baseline} ($a_{\text{mode}}$), which exclusively uses the model's favorites answer, consistently lands around \textbf{0.30}. This performance is very close to the average accuracy result shows in Table \ref{tab:model-evaluation}. This suggests that smaller models rely on a simple way which is always give the preferred answer rather than understanding complex task semantics.

As model size increases, this dependence on a single choice lessens. For instance, the 72B parameter models show a much more uniform choice distribution, with $p_{\text{mode}}$ values dropping below 0.45 and normalized entropy $H_{\text{norm}}$ rising above 0.82. These findings underscore that the apparent choice bias in smaller models is a productive heuristic, allowing them to outperform random chance by exploiting a systematic prior. Acknowledging this artifact is crucial for correctly interpreting the capabilities of low-capacity models

%% file: tables/Bias.tex
\begin{table}[t]
\centering
\renewcommand{\arraystretch}{1.2} 
\setlength{\tabcolsep}{1.5mm}      
\small                          
\begin{tabular}{lcccc}
\toprule
\textbf{Model} & \textbf{$H_{\text{norm}}$} & \textbf{$D_{\text{KL}}$} & \textbf{$p_{\text{mode}}$} & \textbf{$a_{\text{mode}}$} \\
\midrule
InternVL2-4B    & 0.4681 & 0.9337 & 0.6915 & 0.2982 \\
InternVL2-8B    & 0.5587 & 0.9021 & 0.6975 & 0.3063 \\
InternVL2-40B   & 0.7145 & 0.6098 & 0.5786 & 0.3005 \\
InternVL2-76B   & 0.8029 & 0.4650 & 0.4951 & 0.3000 \\
\hline
InternVL3-2B    & 0.5228 & 0.9163 & 0.7197 & 0.3031 \\
InternVL3-8B    & 0.7620 & 0.4981 & 0.5269 & 0.3095 \\
InternVL3-38B   & 0.7776 & 0.5160 & 0.5412 & 0.3009 \\
InternVL3-78B   & 0.8061 & 0.3761 & 0.4961 & 0.3112 \\
\hline
Qwen2.5VL-3B    & 0.5053 & 0.8679 & 0.7063 & 0.3000 \\
Qwen2.5VL-7B    & 0.7020 & 0.6326 & 0.5955 & 0.3067 \\
Qwen2.5VL-32B   & 0.8312 & 0.4417 & 0.4579 & 0.2973 \\
Qwen2.5VL-72B   & 0.8631 & 0.3412 & 0.4359 & 0.3138 \\
\hline
Qwen2VL-2B      & 0.4412 & 1.0985 & 0.7342 & 0.3012 \\
Qwen2VL-7B      & 0.7325 & 0.5981 & 0.5574 & 0.3051 \\
Qwen2VL-72B     & 0.8216 & 0.4126 & 0.4520 & 0.3108 \\
\bottomrule
\end{tabular}
\caption{Average option-bias statistics across several tasks, including Heading Estimation and Map Selection, for various LMMs. We report the normalized entropy ($H_{\text{norm}}$), KL divergence from uniform ($D_{\text{KL}}$), the frequency of the most-predicted answer ($p_{\text{mode}}$), and the accuracy achievable by only guessing that single answer ($a_{\text{mode}}$).}
\label{tab:option-bias}
\end{table}

%% file: secs/5-Conclusion.tex
\section{Conclusion}
\label{sec5}

In this work, we introduced \textbf{GeoX-Bench}, a large-scale benchmark for evaluating cross-view geo-localization and pose estimation in large multimodal models (LMMs). With over 10{,}000 aligned satellite--ground pairs and 750k QA across seven tasks, GeoX-Bench exposes spatial reasoning challenges that prior benchmarks cannot capture.

Our evaluation of more than 25 LMMs reveals a clear gap between coarse geo-localization and fine-grained pose estimation. While many high-parameter models exceed 70\% accuracy on localization tasks, they consistently fail to surpass 30\% on pose estimation as well as intra-map localization which require instance level recognition for both satellite image and the ground image which is a gap not solved by simply scaling parameters.This limitation persists even after general-purpose instruction tuning, suggesting a core architectural weakness. Although geometric instruction tuning offers notable improvements, it does not fully bridge the gap.  Furthermore, we find the performance of smaller models is often artificially inflated by hard-coded biases, confirming that robust, cross-view geometric reasoning remains a significant, unsolved challenge for current LMMs.

Overall, our findings show that current LMMs can perform coarse localization but struggle with precise orientation and cross-view geometric reasoning. \textbf{GeoX-Bench} thus serves as a critical benchmark for driving future progress in geometric intelligence, particularly for safety-critical applications such as autonomous navigation and embodied AI.

%% file: secs/7-ack.tex
\section{Acknowmenge}
This work was supported by the National Natural Science Foundation of China (Grants 62225112, 62301310, 62401365, 62271312, 62132006, and U24A20220), the Sichuan Science and Technology Program (Grant 2024NSFSC1426), and the China Postdoctoral Science Foundation (Grants BX20250411 and 2025M773473).
